\documentclass[9pt,technote,compsoc]{IEEEtran}
\ifCLASSOPTIONcompsoc
  \usepackage[nocompress]{cite}
\else
  \usepackage{cite}
\fi
\ifCLASSINFOpdf
\else
\fi

\usepackage{amsmath,graphicx}
\usepackage{caption}
\usepackage{subcaption}
\usepackage{placeins}
\usepackage{hyperref}
\usepackage{placeins}
\usepackage[boxruled]{algorithm2e}
\usepackage{arydshln}
\usepackage{multirow}
\usepackage{url}
\begin{document}
%
\title{Support-BERT: Predicting Quality of Question-Answer Pairs in MSDN using Deep Bidirectional Transformer}
%
%
%
%

\author{Bhaskar~Sen,
        Nikhil~Gopal,
        and~Xinwei~Xue
\IEEEcompsocitemizethanks{\IEEEcompsocthanksitem B. Sen is with Department
of Electrical and Computer Engineering, University of Minnesota Twin Cities, Minneapolis, MN, 55455. This work was completed during his 2019 appointment at the Microsoft Corporation, One Microsoft Way, Redmond, WA, 98052. \protect\\
\IEEEcompsocthanksitem N. Gopal and X. Xue are with Microsoft Corporation, One Microsoft Way, Redmond, WA, 98052. Correspondence: nigopal@microsoft.com}
}

%
%

\markboth{Arxiv,~Vol.~xx, No.~xx, December~2019}%
{Sen \MakeLowercase{\textit{et al.}}: Support-BERT: Predicting Quality of Question-Answer Pairs in MSDN using Deep Bidirectional Transformer}
%



\IEEEtitleabstractindextext{%
\begin{abstract}
  Quality of questions and answers from community support websites (e.g. Microsoft Developers Network, Stackoverflow, Github, etc.) is difficult to define and a prediction model of quality questions and answers is even more challenging to implement. Previous works have addressed the question quality models and answer quality models separately using meta-features like number of up-votes, trustworthiness of the person posting the questions or answers, titles of the post, and context naive natural language processing features. However, there is a lack of an integrated question-answer quality model for community question answering websites in the literature. In this brief paper, we tackle the quality Q\&A modeling problems from the community support websites using a recently developed deep learning model using bidirectional transformers. We investigate the applicability of transfer learning on Q\&A quality modeling using Bidirectional Encoder Representations from
Transformers (BERT) trained on a separate tasks originally using Wikipedia. It is found that a further pre-training of BERT model along with finetuning on the Q\&As extracted from Microsoft Developer Network (MSDN) can boost the performance of automated quality prediction to more than $80\%$. Furthermore, the implementations are carried out for deploying the finetuned model in real-time scenario using AzureML in Azure knowledge base system.

\end{abstract}

}

\maketitle

\IEEEdisplaynontitleabstractindextext

%
\IEEEpeerreviewmaketitle


%
%
%
%
\vspace{0.5cm}
\ifCLASSOPTIONcompsoc
\IEEEraisesectionheading{\section{Introduction}\label{sec:introduction}\vspace{-0.2cm}}
\else
\fi
\IEEEPARstart{C}ommunity question answering (CQA) websites ({\em e.g Stackoverflow, Github}) have become quite popular   for immediate brief answers of a given question~\cite{vasilescu2013stackoverflow}. Software developers, architects and data scientists regularly visit the relevant forums and websites, on a day-to-day basis for referencing necessary technical contents. In addition, they often use the modified versions of code snippets from the CQA websites for solving their use cases. Hence maintaining high quality answers in those community websites is imperative for their continued relevance in the developers' community.  A common scenario for many questions in the community forums is that there are likely more than one answers for the given question~\cite{shah2010evaluating, liu2008understanding}. However, out of all the available answers, only a few of them are worthwhile in terms of technical quality and usefulness. Finding those quality answers for given questions manually is time consuming, and typically requires a community support engineer (domain expert) to read the answers and record the optimal answer (under the criteria of clarity, technical content, and structure). In addition, a standardized definition of a "high-quality" answer on CQA websites does not exist. Thus, a system that can model high-quality answers based on their technical content, without having to be explicitly defined, is greatly desired in order to circumvent these challenges.  

 However, there are relatively lesser amount of research on understanding what a good quality question-answer pair is for websites like Stackoverflow or Microsoft Developer Network (MSDN) compared to generic CQA websites like Quora or Yahoo!~\cite{shen2015question}. One of the reasons for this is caused by the the excessive technical nature of the contents in those forums. It might be impractical to predict the question-answer quality only based on language semantics. Hence, incorporating technical semantics and content have the potential to improve the answer quality modeling for those forums. 
  
The CQA websites hold a treasure of technical contents which is a database of useful technical questions and corresponding answers on various topics. It can be exploited to further improve their functionality~\cite{pal2012evolution}. Viewing from the machine learning perspective, the CQA websites' language intensive question-answering datasets are rich in resources for automated Q\&A modeling. Specifically recent developments in the natural language processing (NLP) space regarding learning context based representation techniques holds the promise to spearhead the field of automated question-answer quality models~\cite{tian2013towards}. One may ponder what a good quality Q\&A means $-$ the answers accepted in the CQA forums are likely to point towards ground truths that an automated question-answer model should be able to exploit. Another point of interest is how practical the models are in a real-world production scenario. For example, if the latency time during inference is more that 500 ms, the model is unlikely to produce any tangible benefits for practical application. 

\begin{figure*}[ht]
\vspace{-0.1cm}
    \centering
        \centering
        \includegraphics[width=\textwidth]{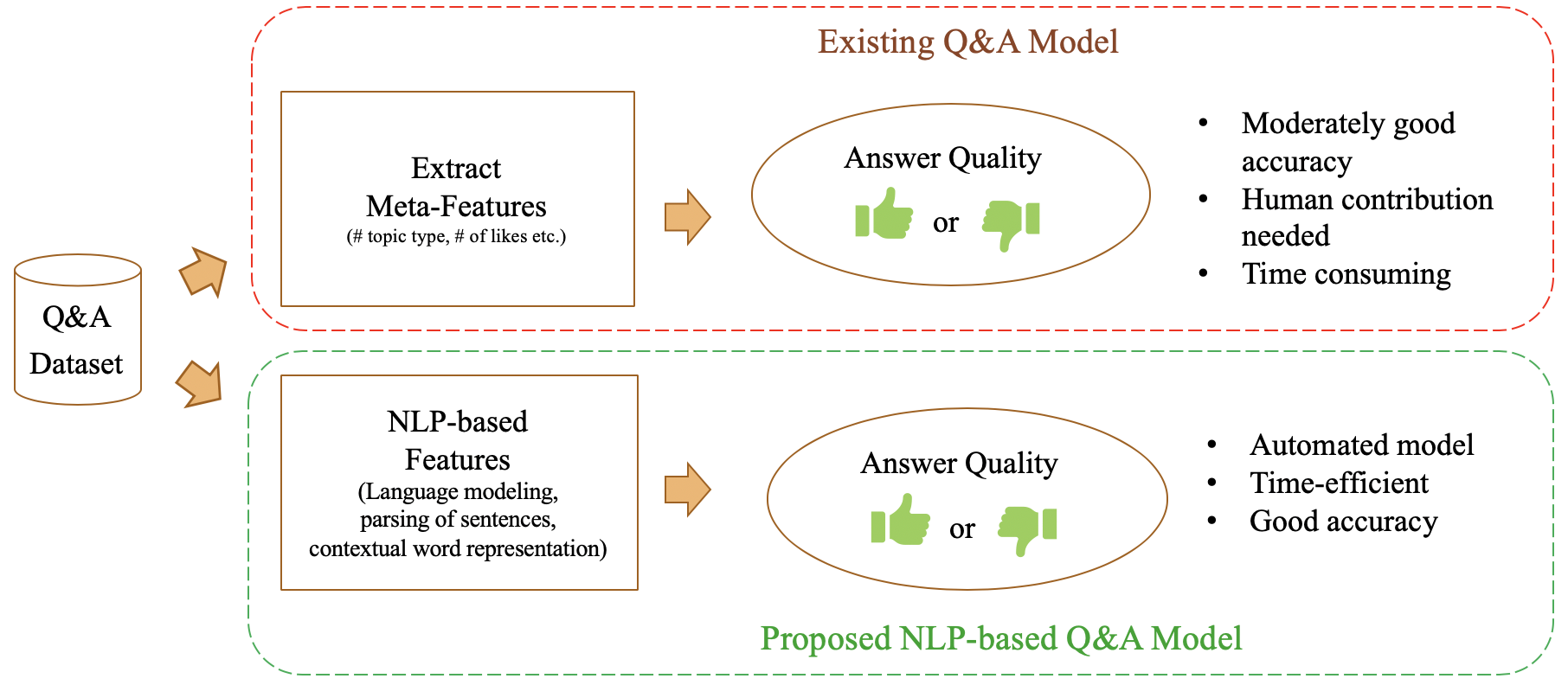}
        \caption{Model with only metafeatures $vs.$ model with only NLP-based features. Existing Q\&A model in Azure knowledge base implements a classifier based meta-features like number of up-votes. The proposed model circumvents this and implements an NLP based classifier.}
    \label{fig:prev_current}
\end{figure*}

In this vein, the work presented in this paper makes substantial progress in both these two dimensions in Q\&A quality modeling. 
\begin{itemize}
    \item First, the paper investigates whether the advancement of NLP techniques in a general setting have tangible benefits in technical content modeling. A whole question-answer pair is investigated for predicting the quality rather than modeling questions and answers separately. Transfer learning~\cite{pan2009survey, tan2018survey} using already pre-trained model using Bidirectional Encoder Representations from Transformers (BERT) is used. The model is again pre-trained and finetuned in the community support space. We call the final model support-BERT. 
    \item Second, the paper shows that a Q\&A quality model with reasonably good performance using a deep neural network can also be implemented within the specified latency. The performance of the model is evidential that our model can be deployed in real-time scenario in Azure knowledge base system.
    
\end{itemize}
 The related pre-training and finetuning code are available from~\footnote{\url{https://github.com/Microsoft/AzureML-BERT}}.

\section{Related Work and Contribution}
Modeling the Q\&A quality in community question answering websites is not new. A number of studies have used different research questions for solving the Q\&A modeling problem. 

\subsection{Predict Good Quality Questions}
Predicting the difficulty of questions was studied in~\cite{hanrahan2012modeling} where they used theory of formal language to create a difficulty level of a technical question from Stackoverflow. Tian {\em et al.}~\cite{tian2013predicting} proposed to solve the quality model by finding best expert users for directing the questions for answer. In addition, modeling quality questions and answers in CQA websites have been well studied.~\cite{ravi2014great} modeled the quality of questions (based on of views and the number of up votes a question has garnered) in Stackoverflow using a topic modeling framework.~\cite{li2011improving} used a recommendation system to find out similar questions from a database. A semi-supervised coupled mutual reinforcement framework was proposed in~\cite{bian2009learning} for simultaneously calculating content quality and user reputation. A number of quality metrics were studied in~\cite{agichtein2008finding} for finiding high quality questions and content. A whole question answering scheme using metafeatures, {\em e.g.,} reputations of co-answerers, relationships between reputation and answer speed, and that the probability of an answer being chosen as the best one, was studied in~\cite{anderson2012discovering}. In contrast to these models, support-BERT only takes the questions and answers as input, and models the quality of them as a pair.

\subsection{Predict Good Quality Answers}

There have been sufficient research for understanding high quality answers for general purpose question answering website like Quora or Yahoo!. Some previous works have extensively focused on understanding question quality, e.g,~\cite{baltadzhieva2015predicting}.  Quality answer prediction has been also studied in~\cite{magnini2002right} using web redundancy information. In addition, classical NLP techniques like textual entailment~\cite{wang2007recognizing}, syntactic features~\cite{grundstrom2014using} and non-textual features, e.g.~\cite{jeon2005finding} have been used to predict answer quality. An ensemble of features were tried for answer qualities in~\cite{tran2015jaist}. Application of deep learning for modeling answer quality is also not new. Attentive neural networks have been applied for answer selection from community websites in~\cite{zhang2017attentive}. Previous studies have also shown that question quality can have a significant impact on the quality of answers received~\cite{agichtein2008finding}. High quality questions can also drive the overall development of the community by attracting more users and fostering knowledge exchange that leads to efficient problem solving.
There has also been work on discovering expert users
in CQA sites, which has mainly focused on modeling expert answerers~\cite{sung2013booming, riahi2012finding}. Work on discovering expert users was often positioned in the context of routing questions to appropriate
answerers (~\cite{li2010routing, li2011question, zhou2012classification}). Our model takes a question-answer pair together and outputs the quality (``accepted'' or ``unaccepted'') without any other meta-features. In addition, the model is structured in a transfer learning framework~\cite{pan2009survey}.


\subsection{Contributions}

We specifically test the following three hypotheses in this paper:
\begin{itemize}
    \item A general purpose language modeling framework (that uses language semantics of Q\&As itself) can be trained to model quality of question and answers.
    \item Incorporating technical semantics of Q\&A structures can model the quality better. 
    \item Although deep learning models may be more accurate for modeling the Q\&A quality, due to the huge number of parameters, it is not efficient to be deployed for online question-answer quality check (or real-time question-answer quality check).
\end{itemize}

The contributions of this paper are as follows:
\begin{itemize}
    \item A state-of-the-art natural language modeling is adopted for modeling technical question-answers from MSDN. To the best of our knowledge, support-BERT is the first domain specific BERT pre-trained on MSDN community data, to transfer technical model semantics from progamming community corpora.
    
    \item The original BERT-medium architecture is utilized for training on the community dataset. We find that transfer learning works surprisingly well in modeling question-answer quality for the language intensive community websites like MSDN.
    
    \item The model was deployed on Azure Kubernetes Service under sub-second latency, {i.e.,} the inference engine is real-time.
    
\end{itemize}
The comparison of the proposed model with respect to traditional machine learning based context-naive answer quality model is demonstrated in Fig.~\ref{fig:prev_current}.

\section{Dataset}
The community Q\&A dataset used in this paper is taken from Microsoft Developer Network~\footnote{\url{https://msdn.microsoft.com/en-us/}}. The dataset consists of a number of meta-features, {\em e.g.}, number of upvotes, reputation of answerers, title of questions, topics that the question-answer pair belongs to, etc. The dataset consists of almost 300,000 Question-Answers, out of which 75,000 are accepted and 225,000 are not accepted. However, note that in our work only texts of the questions and answers are used without any kind of meta-features. The data was minimally preprocessed to remove stopwords, pronoun and participles.  

\section{Method}
In this paper, we use transfer learning for modeling the good quality question-answer pair from MSDN. Specifically the proposed model described below falls within the framework of {\em Inductive self-taught learning}~\cite{pan2009survey}. In natural language processing domain, bidirectional transformers have found a lot of attention recently for their wide-range expressibility and performance in common natural language processing tasks. Transformers have been shown to be effective in many supervised learning tasks where they were trained using different tasks and the learned weights were transferred for finetuning~\cite{devlin2018bert}. Motivated by their wide range of adaption in their state-of-the art performance, we wanted to test if the Bidirectional Encoder Representations from Transformers (BERT) models can model the question-answer pair quality in community space. Two versions of experiments are carried out for the BERT modeling $-$ 1) Finetuning of already pre-trained model and 2) Pre-training + finetuning from the initial check-point. In addition, we experiment with changing a number of vocabularies specific to the MSDN community space and their effect on accuracy. Dataset used in the experiment are taken from publicly available sources $-$ namely Microsoft Developer Network (MSDN). The reason for choosing MSDN is the availability of rich text based technical questions and answers. The methods have also been compared with base-line NLP word representation techniques $-$ TF-IDF~\cite{ramos2003using}, Word2Vec~\cite{goldberg2014word2vec}. The best model from the above experiment was chosen for deployment. Moreover, the model is deployed in Azure Kubernatics Services (AKS) as stand-alone implementation.

\begin{figure*}[ht]
\vspace{-0.1cm}
    \centering
        \centering
        \includegraphics[width=\textwidth]{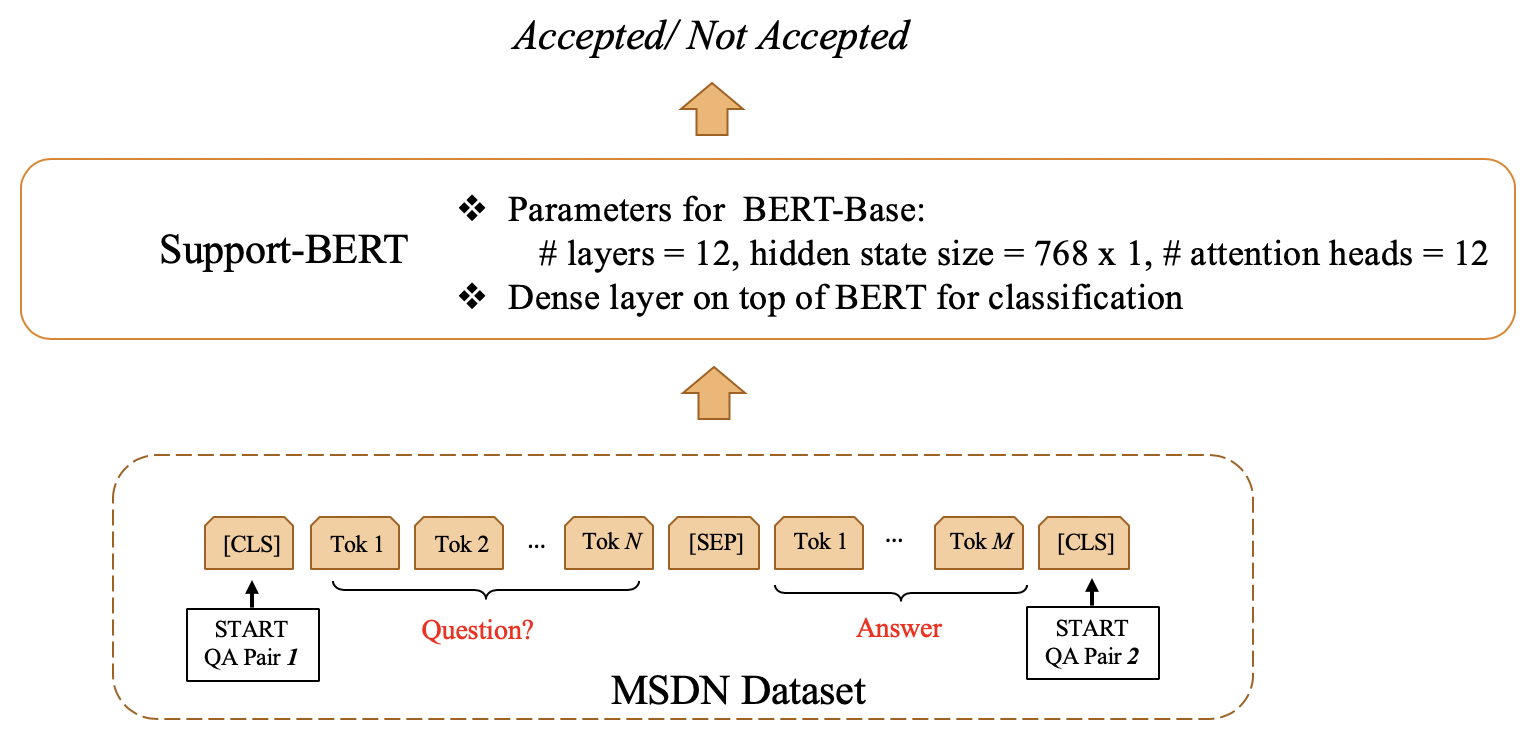}
        \caption{Support-BERT model. A question-answer pair from MSDN is fed to the proposed model, which makes a decision whether the question-answer pair is of good quality.}
    \label{fig:bert_support}
\end{figure*}

\subsection{BERT: Bidirectional Encoder Representation from Transformers}

Using context based word representations for solving natural language processing tasks, e.g., machine translation, question answering and sentence completion have gained popularity in last 10 years. Training on specific NLP tasks (e.g., language modeling~\cite{bengio2003neural}) where word representations were byproducts of the NLP tasks, or direct optimization of word representations based on various hypotheses (\cite{mikolov2013distributed, pennington2014glove}) was conducted to obtain word representations. However, previous studies for representing the words in NLP tasks have mainly utilized representation that are context-naive. More recently, research work on NLP techniques have argued learning context dependent representations. As an example, bi-directional recurrent neural network based language model is used in ELMo~\cite{peters2018deep} that achieved great performance in a number of language tasks. On the other hand, CoVe~\cite{mccann2017learned} makes use of language translation for projecting words into same embedding space based on the context information. The current state-of-the-art in machine translation, multi-task learning for NLP makes use of only attention~\cite{bahdanau2014neural} based neural networks such as transformers~\cite{vaswani2017attention}. BERT~\cite{devlin2018bert} is one such model that exploits the use of contextualized word formats and representation by pre-training the model on a masked language as well as next language prediction framework. Note that previously, because of the uncertainty of NLP models which could not predict the possibility of future words in modeling a context-specific words, bidirectional context-specific models were a combination of left to right and right to left RNN models.  In order to alleviate the problem of extebsive amount of computation required for model bi-directional RNN models, BERT uses a masked word prediction as a task during pre-training, thus removing the constraints of using RNNs. In addition, the model combines next sentence prediction task as well, which encodes context dependent representation for words even in a Q\&A framework. These training criteria on a large text corpus ({\em wikipedia} and {\em bookscorpus}) make BERT model ideal for best preformance on a range of natural language processing task. 

\subsection{BERT as Feature Extractor}
The pre-trained BERT model can be used in transfer learning setting for extracting features in a new domain. In this scenario, Q\&As from community support data is transformed to fixed dimensional vectors using the first few layers of the pre-trained transformer model. The extracted features are then trained and tested using a softmax layer for modeling Q\&A quality.  

\subsection{Finetuning Support-BERT} 
In this experiment, pre-trained BERT model available from tensorflow hub is finetuned without any further pre-training on community support domain. The BERT enoder is appended with a softmax layer and finetuned for 3 iterations for the Q\&A tasks on the MSDN dataset.

\subsection{Pre-Training Support-BERT}

Pre-training a BERT model from scratch is a very slow process. The MSDN dataset, containing technical questions and answers, had a size of around ~300K. In order to fully leverage the technical and language semantics, we started the pre-training from the check-point available from the original BERT model. Then the network was trained for another 100K iterations on the MSDN questions and answers data. In this scenario, masked language modeling was used. The model checkpoints are saved for 20K-100K in 20K iterations progession. This pre-trained network is further finetuned using Q\&A tasks on the MSDN dataset.    

\section{Results and Discussion}
This section describes the results of running the experiments on support-BERT with different configurations. We illustrate and tabulate important results. In addition, we also discuss key observations on the results.

\subsection{BERT as Feature Extractor}
Using BERT as generic feature extractor did not have significant improvement on correctly identifying the quality of Q\&As. Using 50K/50K training set and 50K/50K test set, the maximum accuracy achived on the test set was 0.5340. Using 50K/50K training set and 25K/75K test set, the maximum accuracy achieved on the test set was 0.5890.  

\subsection{Improvement Using Finetuning Support-BERT}
Starting from the checkpoint of BERT model, support-BERT was finetuned for 3 epochs. The finetuning was carried out in supervised learning framework for next sentence prediction. The finetuning for 3 epochs took 3 hours on our machine. There is a visible improvement of the model performance for the prediction task as shown in Table~\ref{table:class1}. In the test scenario for 1:1, accuracy increases up to 0.6966. In the more real-world scenario of 1:3 in test set, the accuracy increases up to 0.7228. 

\begin{table*}[ht]
\centering
\caption{Classification results with BERT+finetuning}
\label{table:class1}
\begin{tabular}{c|c|c|c|c|c|c}
\hline
Training                       & Test                           & \multirow{2}{*}{Accuracy} & \multirow{2}{*}{Precision} & \multirow{2}{*}{Recall} & \multirow{2}{*}{Specificity} &\multirow{2}{*}{F1- Score}\\
\cline{1-2}
Number of accepted/ unaccepted & Number of accepted/ unaccepted &                           &                            &                         &                              \\
\hline
50K/50K                        & 50K/50K                        & 0.6966                    & 0.70                       & 0.6865                  & 0.7125                      &0.6931 \\
50K/50K                        & 25K/75K                        & 0.7228                    & 0.4658                     & 0.7442                  & 0.7156   &0.5729\\
\hline
\end{tabular}
\end{table*}

\subsection{Comparison with Baseline Answer Quality Model}
\begin{figure}[ht]
\vspace{-0.1cm}
    \centering
        \centering
        \includegraphics[width=0.45\textwidth]{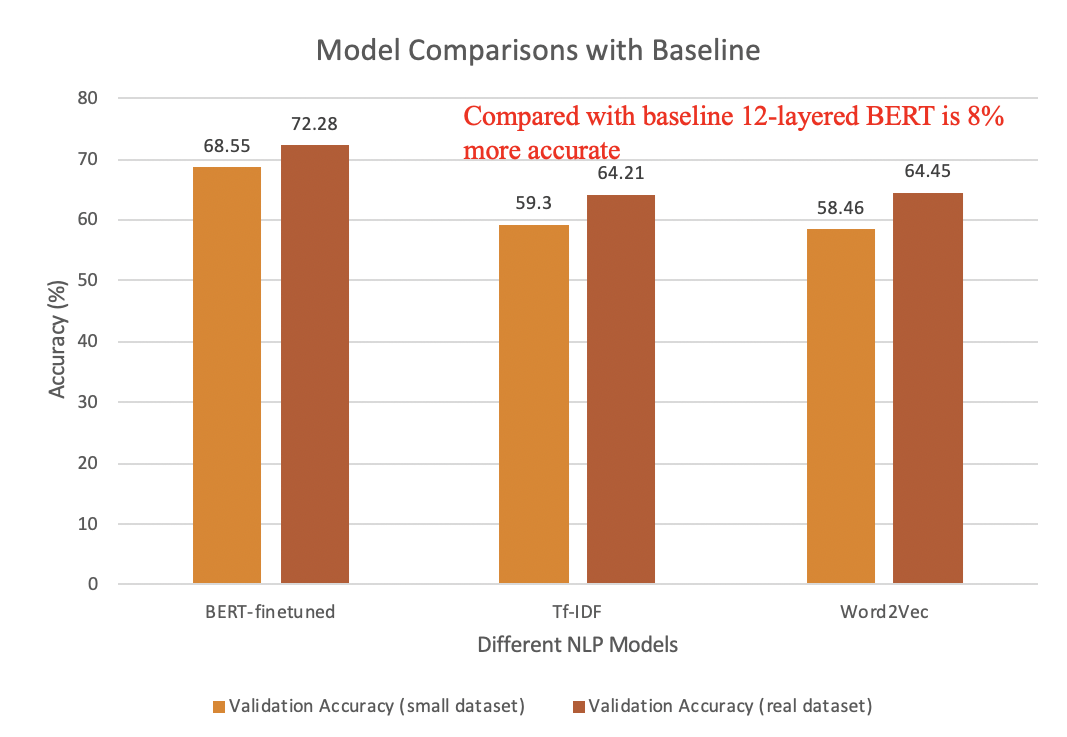}
        \caption{Baseline model accuracy.}
    \label{fig:baseline}
\end{figure}

The proposed support-BERT model with transfer learning was compared with two other baseline models involving context naive language feature, namely TF-IDF and Word2Vec. Both these models performed poorly compared with support-BERT with respect to the Q\&A quality prediction. The results are shown in Fig.~\ref{fig:baseline}.  

\subsection{Adding Domain Specific Words}
\begin{figure*}[ht]
\vspace{-0.1cm}
    \centering
        \centering
        \includegraphics[width=0.95\textwidth]{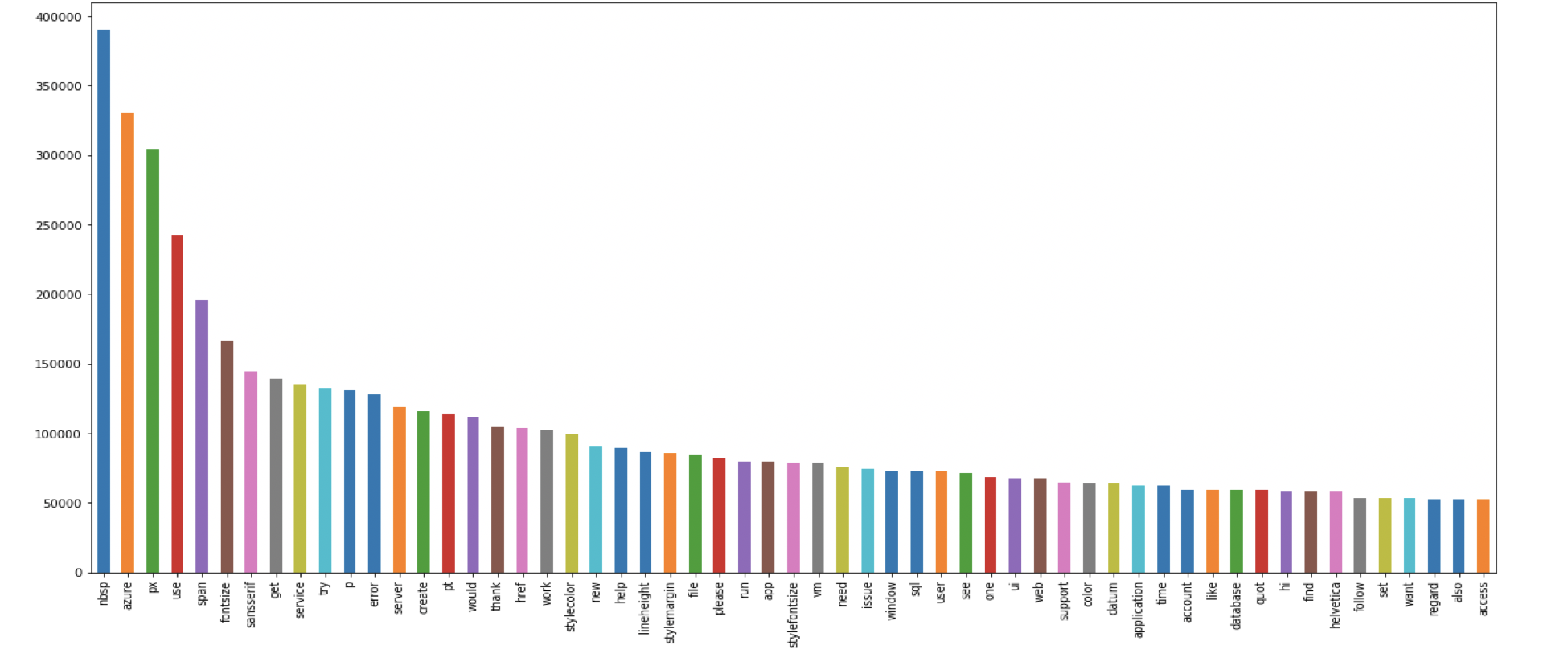}
        \caption{Domain specific word distribution.}
    \label{fig:domainwords}
\end{figure*}

In order to test if the performance of support-BERT is hindered by non-availability of MSDN domain related words, we added top-200 Tf-IDF words from the MSDN corpora to BERT vocabulary. Then the model was further finetuned using the dictionary with added words. However the performance did not improve in this experiment. The accuracy, precision and recall were 0.6865, 0.6957 and 0.6650 respectively. The distribution of top words in the MSDN corpora is shown in Fig.~\ref{fig:domainwords}. 

\subsection{Experiment Regarding Accuracy Drop vs. Number of Layers}
\begin{figure}[ht]
\vspace{-0.1cm}
    \centering
        \centering
        \includegraphics[width=0.45\textwidth]{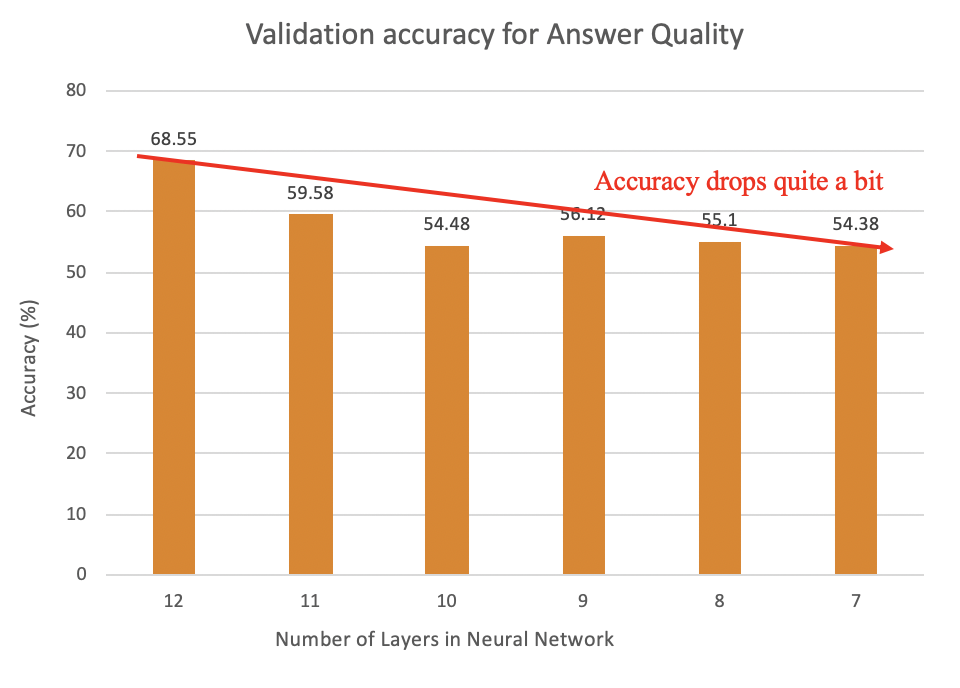}
        \caption{Accuracy $vs.$ number of layers}
    \label{fig:acc_layer}
\end{figure}

Support-BERT has 12-layers of neural network which translates to roughly 110M parameters. In a realtime deployment setting using AzureML, it is possible that the model take a long time for inferencing the quality of Q\&As. In order to understand, the behavior of support-BERT with respect to the number of layers used, we removed the trained layers starting from the last hidden layer. This experiment was carried out on the fined-tuned support BERT as described in Sec. 3.3. The result is illustrated in Fig.~\ref{fig:acc_layer}. The results demonstrate that, removing one layer has a drastic drop in the accuracy values. The accuracy drops by almost 9\%. After that the accuracy drop is less (2\% per layer). 

\begin{figure*}[ht]
\vspace{-0.1cm}
    \centering
        \centering
        \includegraphics[width=\textwidth]{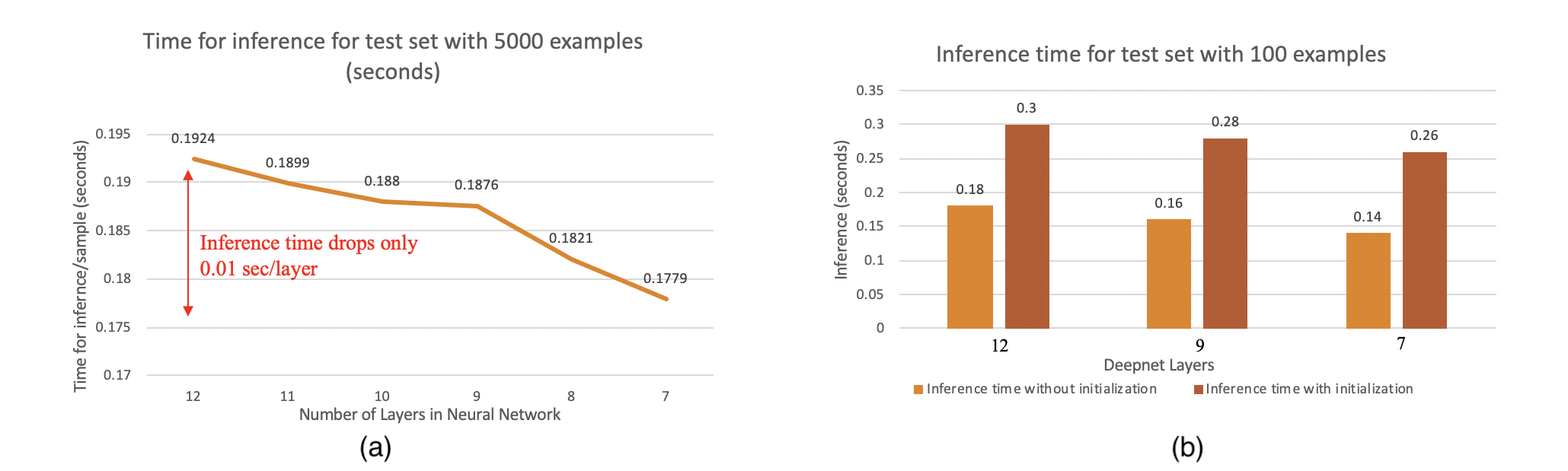}
        \caption{Inference time $vs.$ number of layers. (a) Time for inference/sample $vs.$ number of layers for a test set with 5000 samples. (b) Time for inference/sample $vs.$ number of layers for a test set with 100 samples.}
    \label{fig:infertime_layer}
\end{figure*}

However, the time for inference does not change too much with the removal of layers. In order to investigate the performance of time for inference with respect to number of support-BERT layers, we designed two experiments. The first experiment with test set containing 5K samples measures the total time taken to infer the quality for the batch as in Fig.~\ref{fig:infertime_layer}(a). This involves, retrieving the stored model from disk, initialization of network graph and inference. As a test set is large enough, the effect of initialization is very small per sample. The inference time does not drop too much with lower number of layers in the network. 

The second experiment involves testing with lower number of samples (300 samples in test set). The result is shown in Fig.~\ref{fig:infertime_layer}(b). In this scenario, the effect of initialization during inference is very prominent. Considering the initialization time, per sample inference time is almost 300 ms. However, if we do not consider the initialization time, the per sample inference time is similar to previous experiment.   

\begin{table*}[ht]
\centering
\caption{Classification results with BERT pre-training+finetuning}
\label{table:class2}
\begin{tabular}{c|c|c|c|c|c|c}
\hline
Training                       & Test                           & \multirow{2}{*}{Accuracy} & \multirow{2}{*}{Precision} & \multirow{2}{*}{Recall} & \multirow{2}{*}{Specificity} & \multirow{2}{*}{F-1 Score}\\
\cline{1-2}
Number of accepted/ unaccepted & Number of accepted/ unaccepted &                           &                            &                         &                              \\
\hline
50K/50K                        & 50K/50K                        & 0.8166                    & 0.7768                       & 0.8880                  & 0.7448                 &0.8286      \\
50K/50K                        & 25K/75K                        & 0.7741                    & 0.5279                     & 0.8775                  & 0.7408  &0.6592\\
\hline
\end{tabular}
\end{table*}

\subsection{Improvement Using Pre-training and Finetuning Support-BERT}
Starting from the checkpoint of BERT model trained on Wikipedia in an unsupervised setting, support-BERT was pre-trained on the MSDN community support data for 10 epochs. During pre-training both masked language modeling~\cite{devlin2018bert} and next sentence prediction~\cite{devlin2018bert} framework was used. Note that during pre-training, next sentence prediction involves using sentences defined by words between two full stops. The pre-training for 10 epochs took 48 hours in our machine. Finetuning the model using Q\&As drastically improved the performance. In this case, next sentence prediction model involves using questions as a paragraph (involving more than one actual sentences) as first sentence and answers as a paragraph (involving more than one actual sentences) as next sentence. The results is tabulated in Table~\ref{table:class2}. For test set containing acceptance $vs.$ unaccepted ratio as 1:1, the model was able to identify quality Q\&As 82\% of the time, whereas for test set containing acceptance $vs.$ unaccepted ratio as 1:3, the accuracy value was 0.7741. 

\subsection{Deployment of Support-BERT}
In-house Azure Machine Learning (AzureML) services were used for evidential model deployment process. AzureML is a cloud-based environment that can be used to train, deploy, automate, manage, and track ML models that interoperates with popular open-source tools, such as PyTorch, TensorFlow, and scikit-learn. During our training process, TensorFlow-API for AzureML was used. Support-BERT was trained using Azure NC-6 Virtual Machine (VM), 1 NVIDIA Tesla K80 GPU, 6 vCPU, 56GB MEM, 12GB GPU MEM. The GPUs available in Azure NC VMs are given in Table~\ref{table:gpu}.  We expect that the ``{\em final}" deployment will be done in more advanced GPU and the results are likely to be much faster. 

The {\em winning} model after MSDN domain pre-training and Q\&A specific finetuning was deployed as Azure Container Instances (ACI) on  Azure Kubernetes Service (AKS). We briefly describe the deployment process following~\footnote{\url{https://docs.microsoft.com/en-us/azure/machine-learning/service/how-to-deploy-inferencing-gpus}}. An Azure Machine Learning workspace was created with
a python development environment with the Azure Machine Learning SDK installed. The trained model was registered to the workspace. Specifically, a registered model is a logical container for one or more files that make up the model. For example, if we have a model that's stored in multiple files, we can register them as a single model in the workspace. After registering the files, the model can be downloaded or deployed and the files that was registered can be received. An Azure Kubernate cluster was created with GPU instance for the real-time deployment purpose with NC\_6 GPU VM. For deployment purposes, the procedure given in~\cite{schafer2015class} was followed.  

\begin{figure}[ht]
\vspace{-0.1cm}
    \centering
        \centering
        \includegraphics[width=0.5\textwidth]{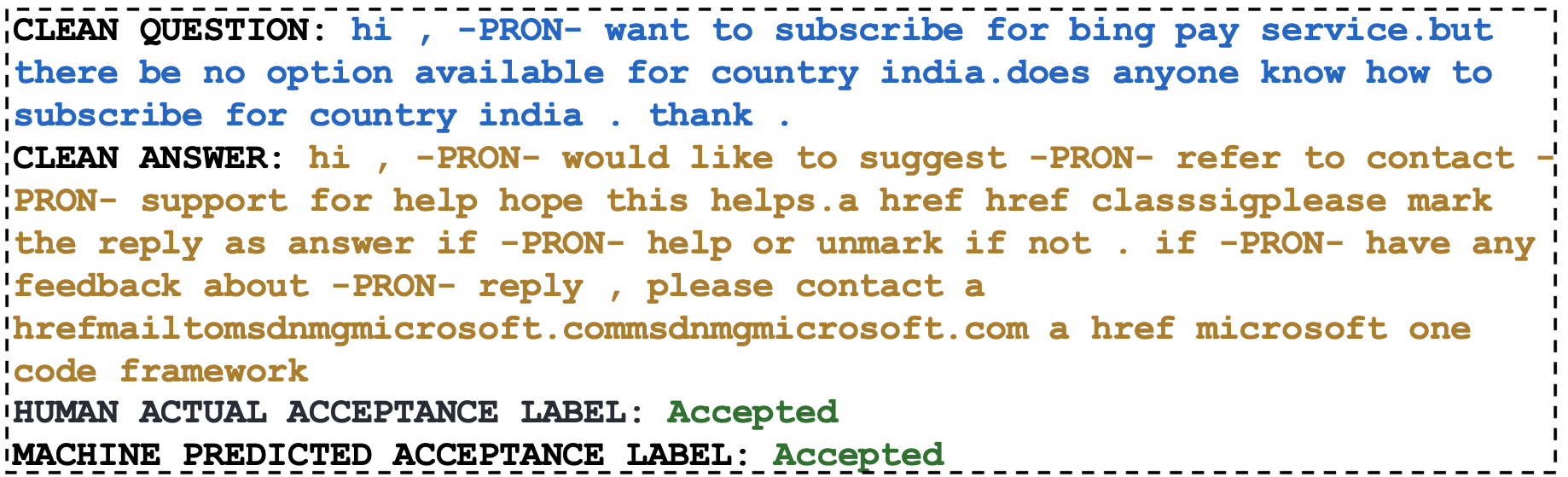}
        \caption{A sample Q\&A.}
    \label{fig:azureml}
\end{figure}

After the deployment, the performance was checked for any degradation on the test data. Once deployed, the latency of new sample query was checked for multiple instances, where the average latency was found to be 110 ms. A sample question and answer from one run of inference from AzureML deployment is shown in Fig.~\ref{fig:azureml}. 

\begin{table*}[ht]
\centering
\caption{GPU configurations available on Azure}
\label{table:gpu}
\begin{tabular}{c|c|c|c|c|c|c|c}
\hline
Size                      & vCPU                           & Memory: GiB & Temp storage (SSD) GiB & 	
GPU & GPU memory: GiB
 &Max data disks &Max NICs                          \\
\hline
Standard\_NC6 &6 &56 &340 &1 &12 &24 &1\\
Standard\_NC12 &12 &112 &680 &2 &24 &48 &2\\
Standard\_NC24 &24 &224 &1440 &4 &48 &64 &4\\
Standard\_NC24r &24 &224 &1440 &4 &48 &64 &4\\
\hline
\end{tabular}
\end{table*}

\section{Conclusion}
In this brief paper, we presented a success of BERT model in CQA support space for modeling good quality question and answers. The proposed support-BERT model after domain specific pre-training and finetuning is an excellent candidate for fast automated decision of the quality Q\&As when a new answer is proposed for a given question. We show that although the goodness of community based CQAs are not well-defined, it is possible to ``mimic" expert validated rules for quality control. In addition, the model proposed in this paper is real-time, thus expediting the process of data analysis to machine learning model implementation step in a tradition data science pipeline. Future work will be directed towards validating the models for other CQA websites like {\em stackoverflow} and {\em github}. In addition, distilling the model to simpler models for inference on a CPU is also of interest. The current finetuned support-BERT model is being evaluated in integration with the Azure knowledge base initiative (providing the high quality relevant answers for questions) to enable support engineers to be more efficient.  


\ifCLASSOPTIONcaptionsoff
\fi
\bibliographystyle{IEEEbib}
\bibliography{strings,refs}

\end{document}